\pdfoutput=1

\documentclass[11pt]{article}

\usepackage[]{ACL2023}

\usepackage{times}
\usepackage{latexsym}

\usepackage[T1]{fontenc}

\usepackage[utf8]{inputenc}

\usepackage{microtype}

\usepackage{inconsolata}
\usepackage[ruled,vlined]{algorithm2e}
\usepackage{algorithmic}
\usepackage{verbatim}
\usepackage{multirow}
\usepackage{threeparttable}
\usepackage{graphicx}
\usepackage{booktabs}
\usepackage{float}
\usepackage{adjustbox}
\usepackage{subfigure}
\usepackage{amssymb}
\usepackage{makecell}
\usepackage{mdwlist}

\title{Clustering-Aware Negative Sampling for Unsupervised Sentence Representation}

\author{ Jinghao Deng\textsuperscript{\rm 1}, Fanqi Wan\textsuperscript{\rm 1}, Tao Yang\textsuperscript{\rm 1},
        Xiaojun Quan\textsuperscript{\rm 1}\thanks{$\;\;$Corresponding authors}, Rui Wang\textsuperscript{\rm 2}
         \\ \textsuperscript{\rm 1}School of Computer Science and Engineering, Sun Yat-sen University, China\\ \textsuperscript{\rm 2}Vipshop (China) Co., Ltd., China \\
         \{dengjh27, wanfq, yangt225\}@mail2.sysu.edu.cn, \\ quanxj3@mail.sysu.edu.cn, mars198356@hotmail.com
}
     
\begin{document}
\maketitle
\begin{abstract}
Contrastive learning has been widely studied in sentence representation learning. However, earlier works mainly focus on the construction of positive examples, while in-batch samples are often simply treated as negative examples. This approach overlooks the importance of selecting appropriate negative examples, potentially leading to a scarcity of hard negatives and the inclusion of false negatives. To address these issues, we propose \textbf{ClusterNS} (\textbf{Cluster}ing-aware \textbf{N}egative \textbf{S}ampling), a novel method that incorporates cluster information into contrastive learning for unsupervised sentence representation learning. We apply a modified K-means clustering algorithm to supply hard negatives and recognize in-batch false negatives during training, aiming to solve the two issues in one unified framework. Experiments on semantic textual similarity (STS) tasks demonstrate that our proposed ClusterNS compares favorably with baselines in unsupervised sentence representation learning. Our code has been made publicly available.\footnote{https://github.com/djz233/ClusterNS}

\end{abstract}

\section{Introduction}
Learning sentence representation is one of the fundamental tasks in natural language processing and has been widely studied \cite{NIPS2015_f442d33f, hill-etal-2016-learning, cer-etal-2018-universal, reimers-gurevych-2019-sentence}. \citet{reimers-gurevych-2019-sentence} show that sentence embeddings produced by BERT \cite{devlin-etal-2019-bert} are even worse than GloVe embeddings \cite{pennington-etal-2014-glove}, attracting more research on sentence representation with pre-trained language models (PLMs) \cite{devlin-etal-2019-bert, liu2019roberta, radford2019language}. \citet{li-etal-2020-sentence} and \citet{ethayarajh-2019-contextual} 
further find out that PLM embeddings suffer from anisotropy, motivating more researchers to study this issue \cite{su2021whitening, gao-etal-2021-simcse}. Besides, \citet{gao-etal-2021-simcse} show that contrastive learning (CL) is able to bring significant improvement to sentence representation. As pointed out by \citet{pmlr-v119-wang20k}, contrastive learning improves the uniformity and alignment of embeddings, thus mitigating the anisotropy issue.

\begin{figure}[t]
    \setlength{\abovecaptionskip}{-0.02cm} 
    \centering
    \includegraphics[width=1.0\linewidth]{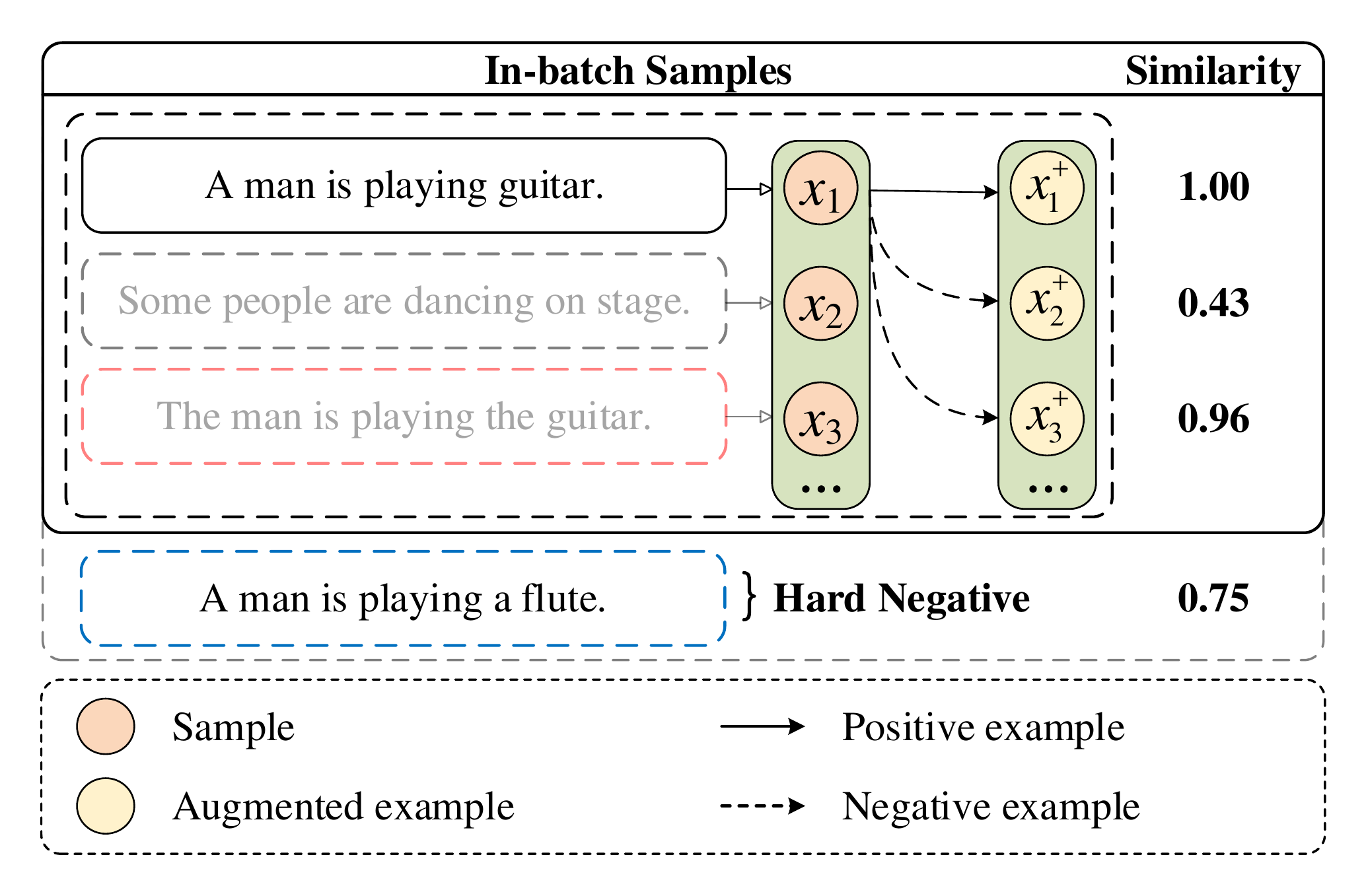}
    \caption{An example of in-batch negatives, a hard negative (in blue dotted box) and a false negative (in red dotted box). Cosine similarity is calculated with SimCSE \protect\cite{gao-etal-2021-simcse}. In-batch negatives may include false negatives, while lacking hard negatives.}
\vspace{-0.6cm}
    \label{fig:neg-example}
\end{figure}

Most previous works of constrastive learning concentrate on the construction of positive examples \cite{kim-etal-2021-self, giorgi-etal-2021-declutr, wu2020clear, yan-etal-2021-consert, gao-etal-2021-simcse, wu-etal-2022-esimcse} and simply treat all other in-batch samples as negatives, which is sub-optimal. 
We show an example in Figure~\ref{fig:neg-example}. 
In this work, we view sentences having higher similarity with the anchor sample as \emph{hard negatives}, which means they are difficult to distinguish from positive samples. When all the negatives are sampled uniformly, the impact of hard negatives is ignored. In addition, various negative samples share different similarity values with the anchor sample and some may be incorrectly labeled (i.e., \textit{false negatives}) and pushed away in the semantic space.

Recently quite a few researchers have demonstrated that hard negatives are important for contrastive learning \cite{Zhang_Zhang_Mensah_Liu_Mao_2022, kalantidis2020hard, xuan2020hard}. However, it is not trivial to obtain enough hard negatives through sampling in the unsupervised learning setting. Admittedly, they can be obtained through retrieval \cite{wang2022improving} or fine-grained data augmentation \cite{wang2022sncse}, but the processes are usually time-consuming. Incorrectly pushing away false negatives in the semantic space is another problem in unsupervised learning scenarios, because all negatives are treated equally. In fact, in-batch negatives are quite diverse in terms of similarity values with the anchor samples. Therefore, false negatives do exist in the batches and auxiliary models may be required to identify them \cite{zhou-etal-2022-debiased}. In sum, we view these issues as the major obstacles to further improve the performance of contrastive learning in unsupervised scenarios.

Since the issues mentioned above have a close connection with similarity, reasonable differentiation of negatives based on similarity is the key. In the meanwhile, clustering is a natural and simple way of grouping samples into various clusters without supervision. Therefore, in this paper, we propose a new negative sampling method called \textbf{ClusterNS} for unsupervised sentence embedding learning, which combines clustering with contrastive learning. 
Specifically, for each mini-batch during training, we cluster them with the K-means algorithm \cite{hartigan1979algorithm}, and for each sample, we select its nearest neighboring centroid (cluster center) as the hard negative. 
Then we treat other sentences belonging to the same cluster as false negatives. Instead of directly taking them as positive samples, we use the Bidirectional Margin Loss to constrain them. Since continuously updating sentence embeddings and the large size of the training dataset pose efficiency challenges for the clustering, we modify the K-means clustering to make it more suitable for training unsupervised sentence representation. 

Overall, our proposed negative sampling approach is simple and easy to be plugged into existing methods, boosting the performance. For example, we improve SimCSE and PromptBERT in $\mathrm{RoBERTa_{base}}$ by 1.41/0.59, and in $\mathrm{BERT_{large}}$ by 0.78/0.88 respectively. The main contributions of this paper are summarized as follows:
\begin{itemize*}
    \item We propose a novel method for unsupervised sentence representation learning, leveraging clustering to solve hard negative and false negative problems in one unified framework.
    \item We modify K-means clustering for unsupervised sentence representation, making it more efficient and achieve better results.
    \item Experiments on STS tasks demonstrate our evident improvement to baselines and we reach 79.74 for $\rm{RoBERTa_{base}}$, the best result with this model.
\end{itemize*}

\section{Related Works}
\subsection{Contrastive Learning}
Contrastive learning is a widely-used method in sentence representation learning. Early works focus on positive examples, and have raised various kinds of effective data augmentations \cite{giorgi-etal-2021-declutr, wu2020clear, yan-etal-2021-consert, gao-etal-2021-simcse}.
Following these works, \citet{wu-etal-2022-esimcse} improve positive construction based on  \citet{gao-etal-2021-simcse}. \citet{zhou-etal-2022-debiased} improve the uniformity  of negative. Besides,  \citet{zhang-etal-2022-contrastive} modify the objective function and \citet{chuang-etal-2022-diffcse} introduce Replaced Token Detection task \cite{Clark2020ELECTRA}, reaching higher performance. 

\begin{figure*}[tbp] 
    \centering
    \includegraphics[width=0.9\linewidth]{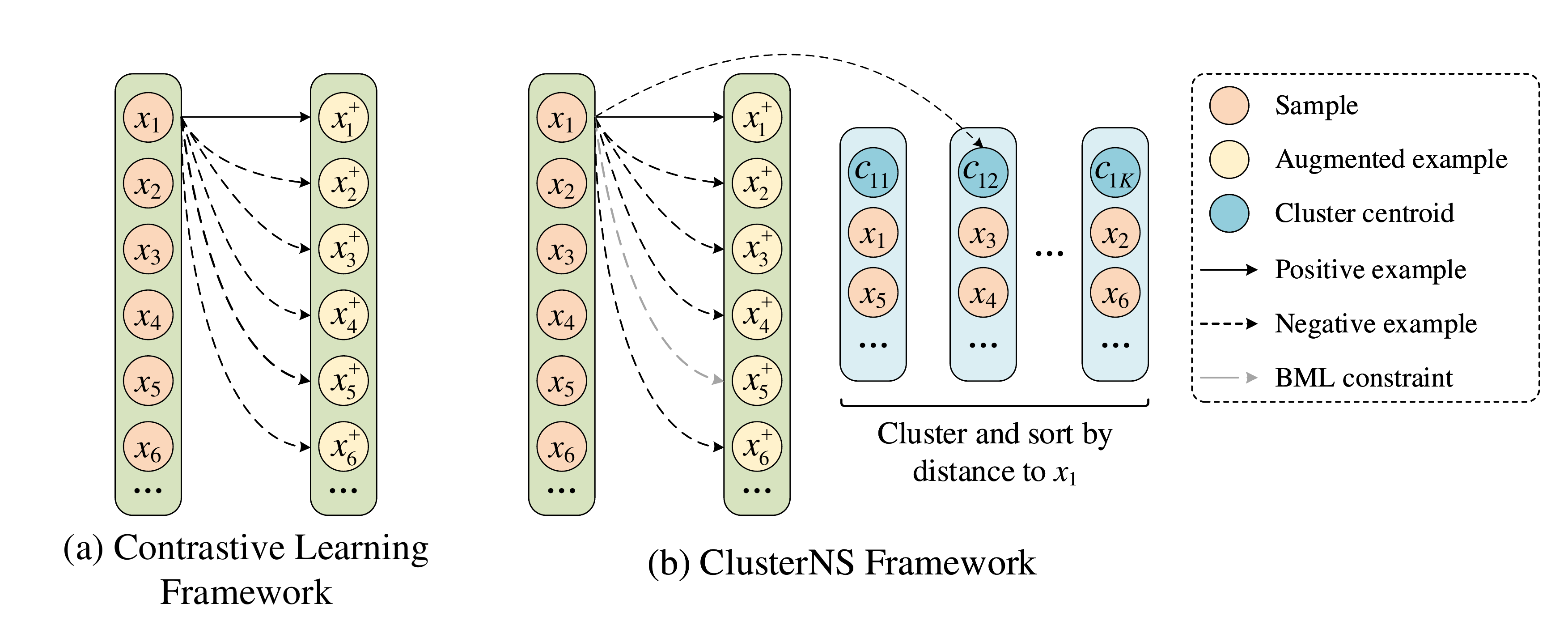}
    \vspace{-0.5cm}
    \caption{Illustrations of the original contrastive learning and our ClusterNS frameworks. Comparing with the naive framework, for the sample $x_{1}$ in the mini-batch, we provide the nearest neighboring centroid $c_{12}$ as the hard negative, and regard the samples in the same cluster (such as $x_{5}$) as false negatives and constrain them by the BML loss.}
    \label{fig:method}
    \vspace{-0.5cm}
\end{figure*}

\subsection{Negative Sampling}
In-batch negative sampling is a common strategy in unsupervised contrastive learning, which may have limitations as we mentioned above. To fix these issues,  
 \citet{Zhang_Zhang_Mensah_Liu_Mao_2022} and  \citet{kalantidis2020hard} synthesize hard negatives by mixing positives with in-batch negatives.  \citet{wang2022sncse} utilize dependency parsing to create the negation of original sentences as soft negatives. Following \citet{jiang2022promptbert} who use different prompt templates as positive,  \citet{zeng2022contrastive} derive negatives from the negation of the templates. The two methods create negatives with fixed templates and rules, thus may introduce bias. \citet{chuang2020debiased} design a debiased contrastive objective that corrects the false negatives without true labels. \citet{zhou-etal-2022-debiased} use a trained model to distinguish false negatives, which results in addition module comparing with our method.
 
\subsection{Neural Clustering}
Clustering methods have been extended to deep learning and used for unsupervised representation learning \cite{xie2016unsupervised, yang2017towards, caron2018deep, li2020prototypical, zhang-etal-2021-supporting}. Prototypical Network \cite{snell2017prototypical}, a variety of clustering, is widely used in few-shot learning \cite{cui-etal-2022-prototypical, ding2020prototypical, gao2019hybrid}. Several works have combined clustering with contrastive learning \cite{li2020prototypical, caron2020unsupervised, zhang-etal-2021-supporting, wang2021pico}. Among them,  \citet{li2020prototypical} argue that clustering encodes high-level semantics, which can augment instance-wise contrastive learning.

\section{Methods}
\label{sec:method}

\subsection{Preliminaries}
\label{sec:simcse}
Our clustering-based negative sampling method for unsupervised sentence representation can be easily integrated with contrastive learning approaches like SimCSE \citep{gao-etal-2021-simcse} or PromptBERT \cite{jiang2022promptbert}. 
An illustration of ClusterNS and the original contrastive learning framework is shown in Figure~\ref{fig:method}. For a sentence $x_i$ in one mini-batch $\{x_i\}^N_{i=1}$ ($N$ samples in each mini-batch), SimCSE uses Dropout \citep{srivastava2014dropout} and PromptBERT uses different prompt-based templates to obtain its positive example $x_i^+$. Then they treat the other samples in the mini-batch as ``default'' negatives and apply the \emph{InfoNCE loss} \citep{oord2018representation} in Eq. (\ref{infonce}),  where $\tau$ is the temperature coefficient. 

\begin{equation} \label{infonce}
    \mathcal{L}_{cl} =  - \log \frac{{{e^{sim\left( {{x_i},x_i^ + } \right)/\tau }}}}{{\sum\limits_{j = 1}^N {{e^{^{sim\left( {{x_i},x_j^ + } \right)/\tau }}}} }}
\end{equation} 

\subsection{Boosting Negative Sampling}
\label{sec:negative method}
Our main contribution of this work is to improve the negative sampling method with clustering. To be more specific, we combine clustering with contrastive learning in the training process, recognizing false negatives in the mini-batch and providing additional hard negatives based on the clustering result. The clustering procedure will be introduced in Section~\ref{sec:K-means method} in detail and for the moment, we assume the samples in each mini-batch have been properly clustered.

Supposing that there are $K$ centroids $c$ = $\{c_i\}^K_{i=1}$ after clustering, standing for $K$ clusters $C$ = $\{C_i\}^K_{i=1}$. For a sample $x_i$ in the mini-batch, we sort the clusters $C$ = [$C_{i1}$,$C_{i2}$,...,$C_{iK}$] and centroids $c$ = [$c_{i1}$,$c_{i2}$,...,$c_{iK}$] by their cosine similarity $cos(x_i, c_{ij})$ with $x_{i}$. In this case, $c_{i1}$ and $c_{iK}$ are the nearest and farthest centroids to $x_i$, respectively. Therefore, $x_i$ is the most similar to $c_{i1}$ and belongs to cluster $C_{i1}$. We define the set $x_{i}^{*}$ = $\{x_{ij}^{*}\}^{count(C_{i1})}_{j=1}$, whose elements belong to $C_{i1}$, the same cluster as $x_i$.
\paragraph{Hard Negatives} 
 \citet{Zhang_Zhang_Mensah_Liu_Mao_2022} show that hard negatives bring stronger gradient signals, which are helpful for further training. The critical question is how to discover or even produce such negatives. In our method, the introduced centroids $c$ can be viewed as hard negative candidates. We get rough groups in mini-batch after clustering and sorting by similarity. For the sample $x_i$, we pick the centroid $c_{i2}$ as its hard negative. The reason is that $c_{i2}$ gets the highest similarity with $x_i$ among all the centroids (except for $c_{i1}$ which $x_i$ belongs to) while having a different cluster. 

In this way, all the samples have proper centroids as their hard negatives, and the training objective $\mathcal{L}_{cl}$ is as follows:

\begin{equation} \label{hn_infonnce}
   \mathcal{L}_{cl} =\!- \log \frac{{{e^{sim\left( {{x_i},x_i^ + } \right)/\tau }}}}{{\sum\limits_{j = 1}^N\!{\left( {{e^{^{sim\left( {{x_i},x_j^ + } \right)/\tau }}}\!+ \mu {e^{^{sim\left( {{x_i},x_j^ - } \right)/\tau }}}} \right)} }}
\end{equation}
where $x_j^-$ is the hard negative corresponding to $x_j$, $\mu$ is the weight of the hard negative. Note that $c_{i1}$ is more similar to $x_i$ compared with $c_{i2}$, which is another candidate for the hard negative. We have compared the different choices in the ablation study described in Section~\ref{sec:ablation}.

\paragraph{False Negatives} For sample $x_i$, we aim to 1) recognize the false negatives in the mini-batch and 2) prevent them from being pushed away incorrectly in the semantic space. For the former, we treat elements in $x_{i}^{*}$ as false negatives, since they belong to the same cluster and share higher similarity with $x_i$.
For the latter, it is unreliable to directly use them as positives, since the labels are missing under the unsupervised setting. 
However, the different similarity between the anchor sample and others can be summarized intuitively as the following Eq. (\ref{neg_relation}):

\begin{equation} \label{neg_relation}
   cos(x_{i}, x_{i}^-) \le cos(x_{i}, x_{ij}^{*}) \le cos(x_{i}, x_{i}^+)
\end{equation}
where $x_{ij}^{*}\in x_{i}^{*}$. False negatives have higher similarity with the anchor than normal negatives while lower similarity than the positives. Inspired by \citet{wang2022sncse}, we introduce the bidirectional margin loss (BML) to model the similarity between the false negative candidates and the anchor:

\begin{equation} \label{delta}
    \Delta_{x_{i}}  = cos(x_{i}, x_{i}^{*})-cos(x_{i}, x_{i}^{+})
\end{equation}
\begin{equation} \label{bml}
    \mathcal{L}_{bml}=ReLU(\Delta_{x_{i}}+\alpha )+ ReLU(-\Delta_{x_{i}}-\beta)
\end{equation}

$\emph{BML loss}$ aims to limit $cos(x_{i}, x_{i}^{*})$ in an appropriate range by limiting $\Delta_{x_{i}}$ in the interval [$-\beta$, $-\alpha$]. Accordingly, we find the potential false negatives in the mini-batch and treat them differently. Combining Eq. (\ref{hn_infonnce}) and Eq. (\ref{bml}), we obtain the final training objective function as follows:
\begin{equation} \label{total_loss}
    \mathcal{L} = \mathcal{L}_{cl} + \lambda \mathcal{L}_{bml}
\end{equation}
where $\lambda$ is a hyperparameter.

\vspace{0.5cm}
\subsection{In-Batch Clustering}
\label{sec:K-means method}
K-means clustering is the base method we use, while we need to overcome computational challenges during the training process. 
It is very inefficient to cluster the large training corpus. However, we need to do clustering frequently due to the continuously updating embeddings. Therefore, we design the training process with clustering in Algorithm~\ref{alg}. Briefly speaking, we use cosine similarity as the distance metric, cluster the mini-batch and update the centroids with momentum at each step. 

\begin{algorithm}[htbp]
	\caption{Training with Clustering.}
	\label{alg}
	\KwIn{Model parameters: $\theta$; Training dataset: $\mathcal{D}$; Total update steps: ${T}$; Warm-up steps: ${S}$}
	\BlankLine
    \begin{algorithmic}[1]
        \FOR{$t = 1$ to $T$}{
        \STATE Get the sentence embeddings $\{x_i\}^N_{i=1}$ for each mini-batch
        
        \IF{${t}$ == ${S}$} 
            \STATE Initialize centroids $c$ with mini-batch samples heuristically
            \ENDIF
        \IF{${t}$ > ${S}$}
        \STATE Update centroids $c$ with $\{x_i\}^N_{i=1}$
        \STATE Provide centroids as hard negatives $\{x_i^-\}^N_{i=1}$ 
        \STATE Calculate $\mathcal{L}_{bml}$ for false negatives
        \ENDIF 
        \STATE Calculate $\mathcal{L}_{cl}$
        \STATE Loss backward and optimize $\theta$
        }

        \ENDFOR
    \end{algorithmic}
\end{algorithm}

\paragraph{Centroids Initialization} We show the initialization in line 3\textendash5 in Algorithm~\ref{alg}. The clustering is not performed at the beginning, since high initial similarity of embeddings harms the performance. Instead, we start clustering 
a few steps after the training starts, being similar to the warm-up process.
When initializing, as line 4 shows, we select $K$ samples as initial centroids heuristically: each centroid to be selected should be the least similar to last centroid.

\begin{table*}[t] 
\renewcommand{\floatpagefraction}{1}
\begin{adjustbox}{width=2.0\columnwidth,center}
\begin{tabular}{lcccccccc}
\hline
Models & STS12 & STS13 & STS14 & STS15 & STS16 & STS-B & SICK-R & Avg. \\ \hline \hline
\multicolumn{9}{c}{\emph{Non-Prompt models}} \\ \hline
GloVe embeddings & 55.14 & 70.66 & 59.73 & 68.25 & 63.66 & 58.02 & 53.76 & 61.32 \\
$\rm{BERT_{base}}$ embeddings & 39.70 & 59.38 & 49.67 & 66.03 & 66.19 & 53.87 & 62.06 & 56.70 \\
$\rm{BERT_{base}}$-flow & 58.40 & 67.10 & 60.85 & 75.16 & 71.22 & 68.66 & 64.47 & 66.55 \\
$\rm{BERT_{base}}$-whitening & 57.83 & 66.90 & 60.90 & 75.08 & 71.31 & 68.24 & 63.73 & 66.28 \\
SimCSE-$\rm{BERT_{base}}$ & 68.40 & 82.41 & 74.38 & 80.91 & 78.56 & 76.85 & \textbf{72.23} & 76.25 \\
{*}ClusterNS-$\rm{BERT_{base}}$ & \textbf{69.93} & \textbf{83.57} & \textbf{76.00} & \textbf{82.44} & \textbf{80.01} & \textbf{78.85} & 72.03 & \textbf{77.55} \\ \hline
$\rm{RoBERTa_{base}}$ embeddings & 32.11 & 56.33 & 45.22 & 61.34 & 61.98 & 54.53 & 62.03 & 53.36 \\
$\rm{RoBERTa_{base}}$-whitening & 46.99 & 63.24 & 57.23 & 71.36 & 68.99 & 61.36 & 62.91 & 61.73 \\
SimCSE-$\rm{RoBERTa_{base}}$ & 70.16 & 81.77 & 73.24 & 81.36 & 80.65 & 80.22 & 68.56 & 76.57 \\
ESimCSE-$\rm{RoBERTa_{base}}$ & 69.90 & 82.50 & 74.68 & 83.19 & 80.30 & 80.99 & \textbf{70.54} & 77.44 \\
DCLR-$\rm{RoBERTa_{base}}$ & 70.01 & 83.08 & 75.09 & \textbf{83.66} & 81.06 & 81.86 & 70.33 & 77.87 \\
{*}ClusterNS-$\rm{RoBERTa_{base}}$ & \textbf{71.17} & \textbf{83.53} & \textbf{75.29} & 82.47 & \textbf{82.25} & \textbf{81.95} & 69.22 & \textbf{77.98} \\ \hline
SimCSE-$\rm{BERT_{large}}$ & 70.88 & 84.16 & 76.43 & 84.50 & 79.76 & 79.26 & 73.88 & 78.41 \\
MixCSE-$\rm{BERT_{large}}$ & \textbf{72.55} & 84.32 & 76.69 & 84.31 & 79.67 & 79.90 & 74.07 & 78.80 \\
DCLR-$\rm{BERT_{large}}$ & 71.87 & 84.83 & 77.37 & \textbf{84.70} & \textbf{79.81} & 79.55 & 74.19 & 78.90 \\
{*}ClusterNS-$\rm{BERT_{large}}$ & 71.64 & \textbf{85.97} & \textbf{77.74} & 83.48 & 79.68 & \textbf{80.80} & \textbf{75.02} & \textbf{79.19} \\ \hline \hline
\multicolumn{9}{c}{\emph{Prompt-based models}} \\ \hline
Prompt$\rm{BERT_{base}}$ & 71.56 & 84.58 & 76.98 & 84.47 & \textbf{80.60} & \textbf{81.60} & 69.87 & 78.54 \\
{*}ClusterNS-$\rm{BERT_{base}}$ & \textbf{72.92} & 84.86 & 77.38 & \textbf{84.52} & 80.23 & 81.58 & 69.53 & 78.72 \\ 
ConPVP-$\rm{BERT_{base}}$ & 71.72 & \textbf{84.95} & \textbf{77.68} & 83.64 & 79.76 & 80.82 & 73.38 & 78.85 \\
SNCSE-$\rm{BERT_{base}}$ & 70.67 & 84.79 & 76.99 & 83.69 & 80.51 & 81.35 & \textbf{74.77} & \textbf{78.97} \\ \hline
Prompt$\rm{RoBERTa_{base}}$ & 73.94 & 84.74 & 77.28 & \textbf{84.99} & 81.74 & 81.88 & 69.50 & 79.15 \\
ConPVP-$\rm{RoBERTa_{base}}$ & 73.20 & 83.22 & 76.24 & 83.37 & 81.49 & 82.18 & \textbf{74.59} & 79.18 \\
SNCSE-$\rm{RoBERTa_{base}}$ & 70.62 & 84.42 & 77.24 & 84.85 & 81.49 & 83.07 & 72.92 & 79.23 \\
{*}ClusterNS-$\rm{RoBERTa_{base}}$ & \textbf{74.02} & \textbf{85.12} & \textbf{77.96} & 84.47 & \textbf{82.84} & \textbf{83.28} & 70.47 & \textbf{79.74} \\ \hline
Prompt$\rm{BERT_{large}}$ & 73.29 & 86.39 & 77.90 & 85.18 & 79.97 & 81.92 & 71.26 & 79.42 \\
ConPVP-$\rm{BERT_{large}}$ & 72.63 & 86.68 & 78.14 & 85.50 & 80.13 & 82.18 & 74.79 & 80.01 \\
SNCSE-$\rm{BERT_{large}}$ & 71.94 & 86.66 & \textbf{78.84} & \textbf{85.74} & 80.72 & 82.29 & \textbf{75.11} & 80.19 \\
{*}ClusterNS-$\rm{BERT_{large}}$ & \textbf{73.99} & \textbf{87.53} & 78.82 & 85.47 & \textbf{80.84} & \textbf{82.85} & 72.59 & \textbf{80.30} \\ \hline
\end{tabular}
\end{adjustbox}
\caption{Overall Results on STS tasks of Spearman’s correlation coefficient. All baseline results are from original or relative papers. We use symbol * to mark our models.  Best results are highlighted in bold. }
\label{tab:main}
\end{table*}

\paragraph{Clustering and Updating} We now describe line 7 in detail. First, we assign each sample into the cluster whose centroid have the highest cosine similarity with the sample. After clustering finishes, we calculate a new centroid embedding for each cluster by averaging embeddings of all samples in the cluster with Eq. (\ref{eq:update}), and then update the centroid in the momentum style with Eq. (\ref{eq:momentum}): 
\begin{equation} \label{eq:update}
    \widetilde{x_i}=\frac{1}{N_i} { \sum_{x_j{\in}C_i}^{} x_j}
\end{equation}
\begin{equation} \label{eq:momentum}
    c_i=(1-\gamma)c_i+\gamma\widetilde{x_i}
\end{equation}
where $\gamma$ is the momentum hyperparameter and $N_i$ indicates the number of elements in cluster $C_i$. Finally, based on the clustering results, we calculate the loss 
and optimize the model step by step (in line 9\textendash12).

The method can be integrated with other contrastive learning models, maintaining high efficiency.


\section{Experiments}

\subsection{Evaluation Setup}
Our experiments are conducted on 7 semantic textual similarity (STS) tasks \cite{semeval-2012-sem,
agirre-etal-2013-sem,
agirre-etal-2014-semeval,
agirre-etal-2015-semeval,
agirre-etal-2016-semeval,
cer-etal-2017-semeval,
marelli-etal-2014-sick} and the models are evaluated with the SentEval Toolkit \cite{conneau-kiela-2018-senteval}. We take the Spearman’s correlation coefficient as the metric and follow the \citet{gao-etal-2021-simcse}'s aggregation method of results.

\subsection{Implementation Details}
\label{implement}
Our code is implemented in Pytorch and Huggingface Transformers. The experiments are run on a single 32G Nvidia Tesla V100 GPU or four 24G Nvidia RTX3090 GPUs. Our models are based on SimCSE \cite{gao-etal-2021-simcse} and PromptBERT \cite{jiang2022promptbert}, and named as \emph{Non-Prompt} ClusterNS and \emph{Prompt-based} ClusterNS, respectively. We use BERT \cite{devlin-etal-2019-bert} and RoBERTa \cite{liu2019roberta} as pre-trained language models for evaluation, with training for 1 epoch and evaluating each 125 steps on the STS-B development set. We also apply early stopping to avoid overfitting. Hyperparameter settings and more training details are listed in Appendix~\ref{sec:hyper}.



\subsection{Main Results}
\label{sec:result}

 We present the experiment results in Table~\ref{tab:main}. We compare with four types of models totally, 1) vanilla embeddings of Glove, BERT and RoBERTa models, we report their results provided by \citet{gao-etal-2021-simcse}. 2) Baseline models: BERT-flow \cite{li-etal-2020-sentence}, BERT-whitening \cite{su2021whitening}, SimCSE \cite{gao-etal-2021-simcse} and PromptBERT \cite{jiang2022promptbert}. 3) SimCSE-based models: MixCSE \cite{Zhang_Zhang_Mensah_Liu_Mao_2022},  DCLR \cite{zhou-etal-2022-debiased} and ESimCSE \cite{wu-etal-2022-esimcse}. 4) PromptBERT-based models: ConPVP \cite{zeng2022contrastive} and SNCSE \cite{wang2022sncse}. We compare SimCSE-based models with \emph{Non-Prompt} ClusterNS, and PromptBERT-based models with \emph{Prompt-based} ClusterNS, respectively. In this way, identical representation of sentence embeddings is guaranteed for a fair comparison.
 
 Our conclusions are as follows, comparing with two baseline models, SimCSE and PromptBERT, all ClusterNS models achieve higher performance, indicating their effectiveness and the importance of negative sampling. 
 For non-prompt models, ClusterNS surpasses MixCSE and DCLR in $\mathrm{BERT_{large}}$, and for prompt-based models, ClusterNS also surpasses ConPVP and SNCSE in $\mathrm{BERT_{large}}$ and $\mathrm{RoBERTa_{base}}$. All these models improve negative samples through sampling or construction, demonstrating our models' strong competitiveness. At last, Prompt-based ClusterNS achieves the state-of-the-art performance of 79.74, which is the best result for models with $\mathrm{RoBERTa_{base}}$.

\subsection{Ablation Study}
\label{sec:ablation}

Our proposed method focuses on two issues, producing hard negatives and processing false negatives. To verify the contributions, we conduct the ablation studies by removing each of the two components on test sets of the STS tasks, with \emph{Non-prompt} BERT and RoBERTa models. As we mentioned in Section~\ref{sec:negative method},  we also replace hard negatives with the most similar centroids to verify our choice of hard negatives (named \emph{repl. harder negative}), 
and replace both centroids for hard and false negatives with random clusters to verify our choice of cluster centroids (named \emph{repl. random clusters}). The results are in Table~\ref{tab:ablation}.

\begin{table}[htbp]
\begin{adjustbox}{width=1.0\columnwidth}
\begin{tabular}{lcc}
\hline
Models & $\rm{BERT_{base}}$ & $\rm{RoBERTa_{base}}$ \\ \hline
ClusterNS & 77.55 & 77.98 \\
\emph{ \quad w/o false negative} & 76.99(-0.56) & 77.83(-0.15) \\
\emph{ \quad w/o hard negative} & 76.03(-1.52)  & 77.22(-0.76) \\ 
\emph{repl. harder negative} & 76.97(-0.58) & 77.84(-0.14) \\ 
\emph{repl. random clusters} & 76.77(-0.78) & 77.85(-0.13) \\ \hline
SimCSE & 76.25 & 76.57 \\ \hline
\end{tabular}
\end{adjustbox}
\caption{Ablation results of our methods (\emph{Non-prompt} Models) on the test set of STS tasks.}
\label{tab:ablation}
\end{table}

We observe from Table~\ref{tab:ablation} that removing either component or replacing any part of models lead to inferior performance: 1) Providing hard negatives yields more improvement, since we create high similarity sample leveraging clustering. 2) Processing false negatives solely (without hard negatives) even further harm the performance, indicating that providing virtual hard negatives is much easier than distinguishing real false negatives. 3) Replacing hard negative with most similar centroids also degrades the performance. Since they belong to the identical cluster, the candidate hard negatives could be actually positive samples. And 4) random clusters are also worse, indicating that the selection of clusters does matter. 
We discuss more hyperparameter settings in Appendix~\ref{sec:supple}. 


\section{Analysis}
\label{sec:discuss}

To obtain more insights about how clustering helps the training process, we visualize the variation of diverse sentence pairs similarity during training after clustering initialization in the \emph{Non-Prompt} ClusterNS-$\rm{RoBERTa_{base}}$ model, and analyze the results in detail.

\subsection{In-Batch Similarity}
\label{sec:in-batch}

\begin{figure}[htbp]
    \centering
    \setlength{\abovecaptionskip}{-0.02cm} 
    \includegraphics[width=0.75\linewidth]{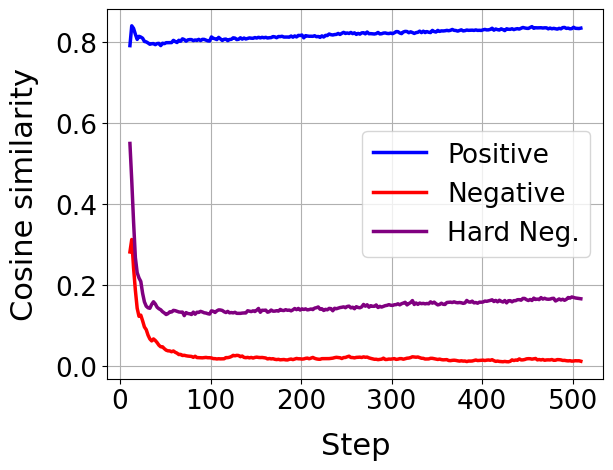}
    \caption{Variation for similarity of positive, in-batch negative and hard negative (Hard Neg.) pairs.}
    \vspace{-0.02cm}
    \label{fig:in-batch}
\end{figure}

We visualize the average similarity of positive, in-batch negative and hard negative sentence pairs in Figure~\ref{fig:in-batch}. We observe that similarity of in-batch negative drops rapidly as training progresses, indicating that in-batch negatives are difficult to provide gradient signal. The hard negatives provided by our method maintain higher similarity, which properly handles the issue. Also notice that the similarity of hard negatives is still much smaller than positive pairs, which avoids confusing the model. 

\begin{figure}[htbp]
    \centering
    \setlength{\abovecaptionskip}{-0.02cm} 
    \includegraphics[width=0.75\linewidth]{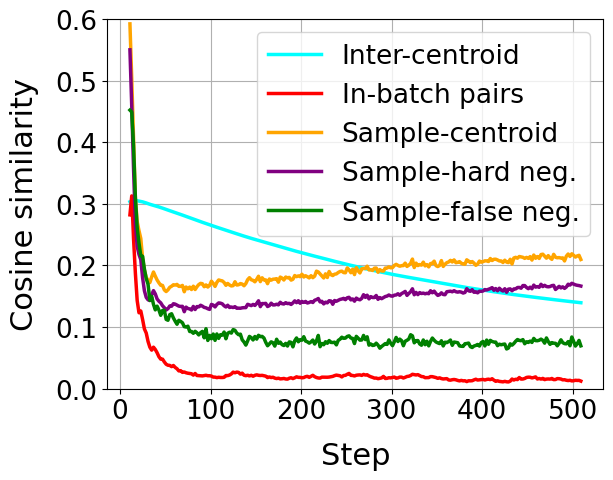}
    \caption{Variation for similarity of sample-nearest centroid pairs (Sample-centroid), Sample-hard negative pairs, Sample-false negative pairs (Intra-cluster member pairs), Inter-centroid pairs and In-batch negative pairs.}
    \label{fig:cluster}
\end{figure}

\subsection{Clustering Similarity}
\label{sec:cluster-sim}

Furthermore, we also visualize the similarity related to clustering. In Figure~\ref{fig:cluster}, we show the average similarity of sample-nearest centroid pairs, sample-hard negative pairs (second nearest centroids, same as hard negative sentence pairs in Figure~\ref{fig:in-batch}), inter-centroid pairs, intra-cluster member pairs (same as false negative pairs) and in-batch negative pairs. First, similarity of inter-centroid pairs decreases during training, demonstrating that clusters representing diverse semantics slowly scatter. Second, false negative pairs get much higher similarity than in-batch negatives, which indicates the importance of recognizing them and the necessity of treating them differently. 
At last, sample-nearest centroid pairs and sample-hard negative pairs maintain high similarity, demonstrating the stability of clustering during the training process.

To answer the question what is a \emph{good} hard negative, we experiment with different similarity levels. 
 We define a symbol $\sigma$, the average similarity threshold of in-batch sentence pairs when the centroids initialize. Since the similarity of hard negative pairs depends on 
 $\sigma$, we adjust the similarity level with various $\sigma$ settings.
 
\begin{figure}[htbp]
    \centering
    \setlength{\abovecaptionskip}{-0.01cm} 
    \includegraphics[width=0.8\linewidth]{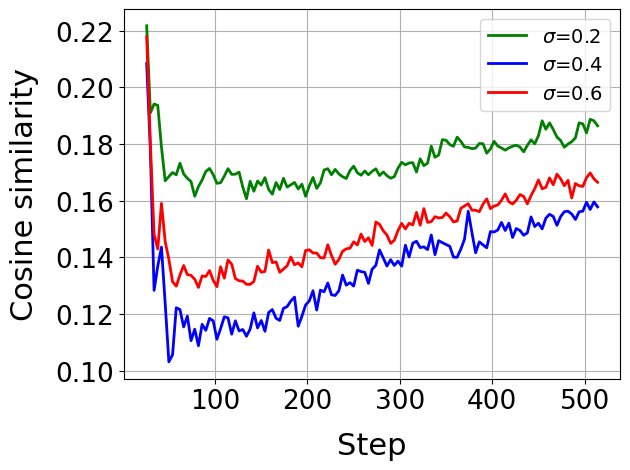}
    \caption{Varition for average similarity of hard negative pairs in different threshold $\sigma$}
    \label{fig:sigma}
    \vspace{-0.2cm}
\end{figure}

 We show the results in Figure~\ref{fig:sigma} and Table~\ref{tab:sigma}. As we set the threshold $\sigma$ smaller, clustering begins later and hard negatives gets larger similarity (with the anchor sample), meaning that starting clustering too early leads to less optimal hard negative candidates. The best performance is achieved at $\sigma$ = 0.4, the middle similarity level, verifying the finding in our ablation study, i.e., hard negatives are not the most similar samples.

\begin{table}[htbp]
\centering
\resizebox{0.8\linewidth}{!}{
\begin{tabular}{c|ccc}
\hline
$\sigma$ & 0.2 & 0.4 & 0.6 \\ \hline
Similarity & 0.1814 & 0.1572 & 0.1442 \\ 
Avg. STS & 77.02 & \textbf{77.98} & 77.30 \\ \hline
$\sigma$ wo. $L_{bml}$ & 0.2 & 0.4 & 0.6 \\ \hline
Similarity & 0.1797 & 0.1574 & 0.1453 \\
Avg. STS & 77.43 & 77.83 & 77.46 \\ \hline
\end{tabular}
}
\caption{The average results of STS test sets in different threshold $\sigma$ with$\backslash$without $L_{bml}$, similarity means the average similarity of hard negative pairs at the last checkpoints.}
\label{tab:sigma}
\vspace{-0.5cm}
\end{table}

 In Figure~\ref{fig:cluster}, similarity of false negative pairs is much smaller comparing with positive pairs, which shows the distinction between positive and false negative samples. 
False negatives are usually regarded as positive samples in supervised learning, while it is difficult to recognize precisely in the unsupervised setting. We argue that false negatives retrieved by our methods share similar topics with the anchors, leading to higher similarity than ``normal'' negatives and lower similarity than positives. We use Eq. (\ref{bml}) to constrain false negatives based on this hypothesis. We also do case studies and experiments to approve it in Appendix~\ref{sec:case}.

To verify our choice of the BML loss, we implement experiments on different processing strategies of false negatives. We compare BML loss with two common strategies: use false negatives as positives and mask all the false negatives. Results in Table~\ref{tab:bml_ablation} demonstrate the superiority of the BML loss. 

\begin{table}[htbp]
\centering
\resizebox{0.75\linewidth}{!}{
\begin{tabular}{lc}
\hline
Models & Avg. STS \\ \hline
ClusterNS & 77.98 \\
\textit{ \quad w/o BML loss} & 77.83 \\
\textit{Mask all false negatives} & 77.40 \\
\textit{Use as positives} & 42.33 \\ \hline
\end{tabular}
}
\caption{ Comparison of different false negative processing on STS test sets. \emph{w/o BML loss} is the same as \emph{w/o false negative} in Table \ref{tab:ablation}.}
\label{tab:bml_ablation}
\vspace{-0.5cm}
\end{table}

\begin{table}[htbp]
\begin{tabular}{l|ccc}
\hline
Models & RoBERTa & SimCSE & ClusterNS \\ \hline
AMI & 0.6926 & 0.7078 & 0.7355 \\ \hline
\end{tabular}
\caption{AMI score for K-means clustering (K=14) on DBpedia dataset. We use \emph{Non-Prompt} ClusterNS for comparison. Higher values are better.}
\label{tab:ami}
\vspace{-0.55cm}
\end{table}

\begin{table*}[tbp]
\centering
\begin{adjustbox}{width=2\columnwidth,center}
\begin{tabular}{lccccccccc}
\hline
Models & AG & Bio & Go-S & G-T & G-TS & SS & SO & Tweet & Avg. \\ \hline
SimCSE-$\rm{BERT_{base}}$ & 74.46 & 35.64 & 59.01 & 57.92 & 64.18 & 67.09 & 50.78 & \textbf{54.71} & 57.97 \\
ClusterNS-$\rm{BERT_{base}}$ & \textbf{77.38} & \textbf{37.29} & \textbf{61.69} & \textbf{59.37} & \textbf{66.47} & \textbf{69.65} & \textbf{72.92} & 53.71 & \textbf{62.31} \\ \hline
SimCSE-$\rm{RoBERTa_{base}}$ & \textbf{69.71} & \textbf{37.35} & \textbf{60.89} & 57.66 & 65.05 & 46.90 & 69.00 & \textbf{51.89} & 57.31 \\
ClusterNS-$\rm{RoBERTa_{base}}$ & 65.00 & 36.38 & 58.58 & \textbf{57.88} & \textbf{65.54} & \textbf{52.55} & \textbf{74.38} & 51.63 & \textbf{57.74} \\ \hline
\end{tabular}
\end{adjustbox}
\caption{Clustering accuracy on short text clustering datasets. We use \emph{Non-Prompt} ClusterNS for comparison and evaluate on $\mathrm{BERT_{base}}$ and $\mathrm{RoBERTa_{base}}$. We reproduce all baseline results based on provided checkpoints. Best results are highlighted in bold.}
\label{tab:short-text}
\end{table*}

\section{Clustering Evaluation}
\label{sec: cluster-eval}
We also evaluate the quality of sentence embedding through clustering. We first use the DBpedia dataset \cite{brummer-etal-2016-dbpedia}, an ontology classification dataset extracted from Wikipedia and consists of 14 classes in total. We implement K-means clustering (\emph{K}=14) on the sentence embeddings of DBpedia, and take the adjusted mutual information (AMI) score as the evaluation metric following \citet{li2020prototypical}. The results in Table~\ref{tab:ami}\ show that both sentence embedding models improve the AMI score, indicating that the cluster performance is positively correlated with the quality of sentence embeddings. ClusterNS achieves a higher AMI score than SimCSE, verifying its effectiveness. 

Furthermore, we follow \citet{zhang-etal-2021-pairwise} to conduct a more comprehensive evaluation of the short text clustering on 8 datasets \footnote{https://github.com/rashadulrakib/short-text-clustering-enhancement}, including AgNews (AG) \cite{zhang2015text}, Biomedical (Bio) \cite{xu2017self}, SearchSnippets (SS) \cite{phan2008learning}, StackOverflow (SO) \cite{xu2017self}, GoogleNews (G-T, G-S, G-TS) \cite{yin2016model} and Tweet \cite{yin2016model}. We perform K-means clustering on the sentence embeddings and take the clustering accuracy as the evaluation metric. Results are shown in Table \ref{tab:short-text}. Our ClusterNS models achieve higher performance than SimCSE in both two models, with an overall improvement of 4.34 in $\mathrm{BERT_{base}}$. Both main experiments and two clustering evaluations show the improvement of our method to the baseline, and verify the effectiveness of improved negative sampling. More details about evaluation metrics are shown in Appendix \ref{sec:apd-cluster}.

\section{Alignment and Uniformity}
\label{sec: U and A}

To investigate how ClusterNS improves the sentence embedding, we conduct further analyses on two widely used metrics in contrastive learning proposed by \citet{pmlr-v119-wang20k}, \emph{alignment} and \emph{uniformity}. Alignment measures the expected distance between the embeddings of positive pairs:

\begin{equation}
    \label{eq:align}
    \mathcal{L}_{align} \triangleq  \mathop{\mathbb{E}}_{(x,x^+)\sim p_{\mathrm{pos} }} \Vert f(x)-f(x^+) \Vert^2
\end{equation}

And uniformity measures the expected distance between the embeddings of all sentence pairs:
\begin{equation}
    \label{eq:uniform}
    \mathcal{L}_{uniform} \triangleq \log  \mathop{\mathbb{E}}_{(x,y)\sim p_{\mathrm{data} }} 
e^{-2\Vert f(x)-f(y) \Vert^2}
\end{equation}

\begin{figure}[htbp]
    \centering
    \includegraphics[width=1\linewidth]{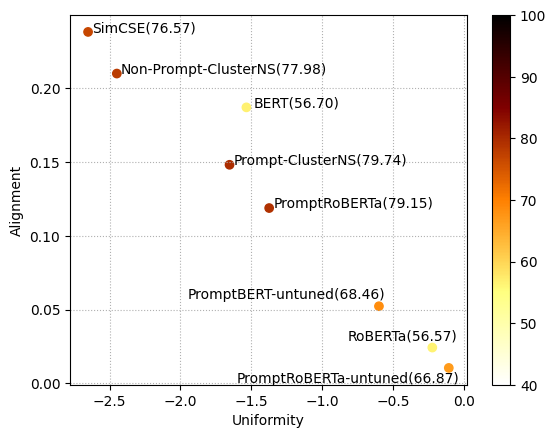}
    \caption{Alignment and uniformity for different sentence
embedding models on the STS-B dataset. \emph{untuned} means the models are not fine-tuned. We mainly use $\mathrm{RoBERTa_{base}}$ models. Lower values are better.}
    \label{fig:au}
    \vspace{-0.5cm}
\end{figure}

Both metrics are better when the numbers are lower. We use the STS-B dataset to calculate the alignment and uniformity, and consider the sentence pairs with score higher than 4 as positive pairs. We show the alignment and uniformity of different models in Figure~\ref{fig:au}, along with the average STS test results. We observe that ClusterNS strikes a balance between alignment and uniformity, improving the weaker metric at the expense of the stronger one to reach a better balance. For the non-prompt models, SimCSE has great uniformity but weaker alignment compared to vanilla BERT and RoBERTa. ClusterNS optimizes the alignment. On the other hand, Prompt-based ClusterNS optimizes the uniformity since PromptRoBERTa performs the opposite of SimCSE. Besides, RoBERTa may suffer server anisotropy than BERT, meaning that sentence embeddings are squeezed in a more crowded part of the semantic space. Therefore, RoBERTa and PromptRoBERTa-untuned have extreme low value of alignment, but poor uniformity. 

\section{Conclusion}
In this paper, we propose ClusterNS, a novel approach that focuses on improving the negative sampling for contrastive learning in unsupervised sentence representation learning. We integrate clustering into the training process and use the clustering results to generate additional hard negatives and identify false negatives for each sample. We also use a bidirectional margin loss to constrain the false negatives. Our experiments on STS tasks show improvements over baseline models and demonstrate the effectiveness of ClusterNS. Through this work, we demonstrate that it is valuable to pay more attention to negative sampling when applying contrastive learning for sentence representation.

\section*{Acknowledgements}
We appreciate the anonymous reviewers for their valuable comments. We thank Zhaoyang Wang for his support. This work was supported by the National Natural Science Foundation of China (No. 62176270), the Guangdong Basic and Applied Basic Research Foundation (No. 2023A1515012832), and the Program for Guangdong Introducing Innovative and Entrepreneurial Teams (No. 2017ZT07X355).

\section*{Limitations}
Our work has two limitations. First, we update the cluster centroids at each step during training, which requires a large mini-batch to maintain clustering accuracy and consumes more GPU memory. Second, our method still may not identify false negatives accurately, as we use the training model for coarse-grained clustering rather than a well-trained model. We leave the improvement of memory consumption and further improving false negative discrimination for the future.

\section*{Ethics Statement}
All datasets used in our work are from public sources, which do not consist private information. We strictly followed the data usage policy. Any research based on our work must sign an ethical statement and ensure that they do not infer user privacy from it.


\bibliography{anthology,custom}
\bibliographystyle{acl_natbib}

\appendix

\section{Training Details}
\label{sec:hyper}

 We do grid search for the hyperparameters and list the searching space below.
\begin{itemize*}
    \item Total batch size [256, 512]
    \item Learning rate [1e-5, 3e-5, 5e-5] 
    \item Hard negative weight $\mu$ [1.0]
    \item Number of cluster $K$ [96, 128, 256]
    \item Momentum $\gamma$ [1e-3, 5e-4, 1e-4]
    \item Similarity threshold $\sigma$ [0.2 0.3 0.4 0.5 0.6]
    \item Weight of $\mathcal{L}_{bml}$ [1e-2, 1e-3, 1e-4, 1e-5]
    \item Upper Bound of $\mathcal{L}_{bml}$ $\alpha$ [0, 0.05, 0.1, 0.15, 0.2, 0.25]
    \item Lower Bound of $\mathcal{L}_{bml}$ $\beta$ [0.3, 0.4, 0.5, 0.6]
\end{itemize*}

Our method has two main improvements on hard negative and false negative, respectively. We apply both improvements to most of the models except one of them. We list the information in detail in Table \ref{tab:hyper}. More hyperparameter experiments are discussed in Appendix \ref{sec:supple}.

\begin{table}[htbp]
\resizebox{\linewidth}{!}{
\begin{tabular}{lcccc}
\hline
\multicolumn{1}{c}{$\emph{Non-Prompt}$} & \multicolumn{2}{c}{BERT} & RoBERTa \\
\multicolumn{1}{c}{$\emph{Models}$} & Base & Large & Base \\ \hline
Hard Negative & $\checkmark$ & $\checkmark$ & $\checkmark$  \\
False Negative & $\checkmark$ & $\checkmark$ & $\checkmark$  \\ \hline
\multicolumn{1}{c}{$\emph{Prompt-based}$} & \multicolumn{2}{c}{BERT} & RoBERTa \\
\multicolumn{1}{c}{$\emph{Models}$} & Base & Large & Base \\ \hline
Hard Negative & $\checkmark$ & $\checkmark$ &   \\
False Negative & $\checkmark$ & $\checkmark$ & $\checkmark$ \\ \hline
\end{tabular}
}
\caption{Hyperparameter settings that whether to apply both improvements on the models.}
\label{tab:hyper}

\end{table}

\begin{table*}[tbp]
\vspace{-0.5cm}
\begin{adjustbox}{width=1.8\columnwidth,center}
\begin{tabular}{lcccccccc}
\hline
Model & MR & CR & SUBJ & MPQA & SST & TREC & MRPC & Avg \\ \hline
GloVe embeddings & 77.25 & 78.30 & 91.17 & 87.85 & 80.18 & 83.00 & 72.87 & 81.52 \\
Avg. BERT embeddings & 78.66 & 86.25 & 94.37 & 88.66 & 84.40 & 92.80 & 69.54 & 84.94 \\
BERT-{[}CLS{]} embedding & 78.68 & 84.85 & 94.21 & 88.23 & 84.13 & 91.40 & 71.13 & 84.66 \\
SimCSE-$\rm{BERT_{base}}$  & 81.18 & 86.46 & 94.45 & 88.88 & 85.50 & 89.80 & 74.43 & \textbf{85.81} \\
w/ MLM & \textbf{82.92} & \textbf{87.23} & \textbf{95.71} & 88.73 & 86.81 & 87.01 & \textbf{78.07} & 86.64 \\
ClusterNS-$\rm{BERT_{base}}$ & 82.01 & 85.46 & 94.44 & \textbf{89.09} & 86.27 & 88.80 & 73.57 & 85.66 \\
w/ MLM & 82.79 & 86.84 & 95.29 & 88.04 & \textbf{86.88} & \textbf{91.80} & 76.99 & \textbf{86.95}  \\ \hline
SimCSE-$\rm{RoBERTa_{base}}$ & 81.04 & 87.74 & 93.28 & 86.94 & 86.60 & 84.60 & 73.68 & 84.84 \\
w/ MLM & 83.37 & 87.76 & \textbf{95.05} & 87.16 & \textbf{89.02} & 90.80 & 75.13 & 86.90 \\
ClusterNS-$\rm{RoBERTa_{base}}$ & 81.78 & 86.65 & 93.21 & \textbf{87.85} & 87.53 & 84.00 & 76.46 & \textbf{85.35} \\
w/ MLM & \textbf{83.51} & \textbf{88.11} & 94.56 & 86.04 & 88.85 & \textbf{92.40} & \textbf{76.70} & \textbf{87.17} \\ \hline
\end{tabular}
\end{adjustbox}
\caption{Transfer task results of different sentence embedding models. Best results are highlighted in bold.}
\label{tab:tr}
\end{table*}

\section{Transfer Tasks}
\label{sec:transfer}

Following previous works, we also evaluate our models on seven transfer tasks: MR \cite{pang-lee-2005-seeing}, CR \cite{hu2004mining}, SUBJ \cite{pang-lee-2004-sentimental}, MPQA \cite{wiebe2005annotating}, SST-2 \cite{socher-etal-2013-recursive}, TREC \cite{voorhees2000building} and MRPC \cite{dolan-brockett-2005-automatically}. We evaluate with \emph{Non-Prompt} ClusterNS models, and use the default configurations in SentEval Toolkit. Results are showed in Table \ref{tab:tr}. Most of our models achieve higher performance than SimCSE and the auxiliary MLM task also works for our methods.

\section{False Negative Details}
\label{sec:case}

We show the case study in Table~\ref{tab:cluster-example}. As we mentioned in Section~\ref{sec:discuss}, our method is able to cluster sentences with similar topics such as religion and music, demonstrating that clustering captures higher-level semantics. However, intra-cluster sentences do not necessarily carry the same meaning and thus they are not suitable to be used as positives directly.

\begin{figure}[htbp]
\setlength{\abovecaptionskip}{-0.01cm} 
    \centering
    \includegraphics[width=0.9\linewidth]{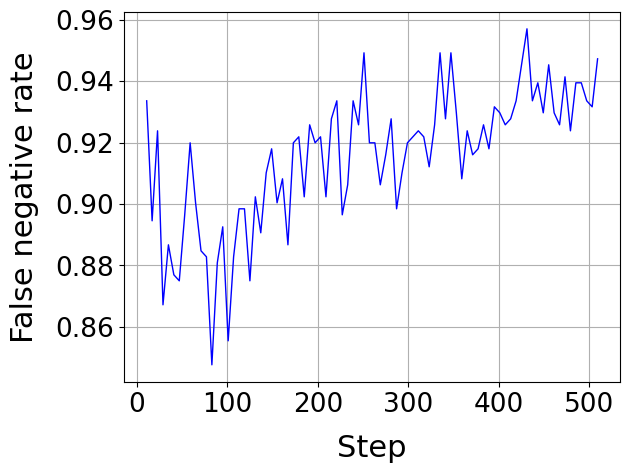}
    \caption{Variation for false negatives rate of \emph{Non-Prompt} ClusterNS-$\mathrm{RoBERTa_{base}}$ in the  training process.}
    \label{fig:fn-rate}
\end{figure}

We also show the variation tendency of the false negative rate in Figure~\ref{fig:fn-rate}, which is equivalent to the sample percentage of clusters having more than two elements (i.e., the intra-cluster members are the false negatives of each other). We observe that the false negative rate maintains a high percentage in the whole training process, which verifies the necessity to specific handling the false negatives.

\section{Clustering Evaluation Details}
\label{sec:apd-cluster}

We use adjusted mutual information (AMI) score or clustering accuracy to evaluate clustering performance. AMI score measures the agreement between ground truth labels and clustering results. Two identical label assignments get the AMI score of 1, and two random label assignments are expected to get AMI score of 0. Clustering accuracy measures the clustering agreement with accuracy metric, which need to map clustering results to ground truth labels with Hungary algorithm in advance. 


\section{Supplement Experiments}
\label{sec:supple}

\subsection{Batch Size and Cluster Number}
\begin{figure}[htbp]
    \centering
    \includegraphics[width=0.8\linewidth]{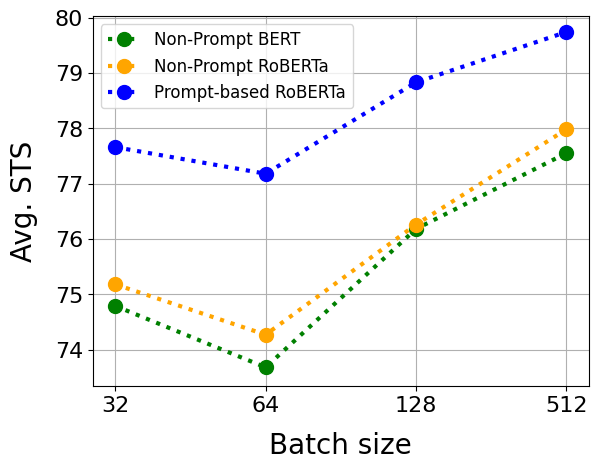}
    \caption{Comparisons of different batch size.}
    \label{fig:bsz}
\end{figure}

\begin{figure}[htbp]
    \centering
    \includegraphics[width=0.8\linewidth]{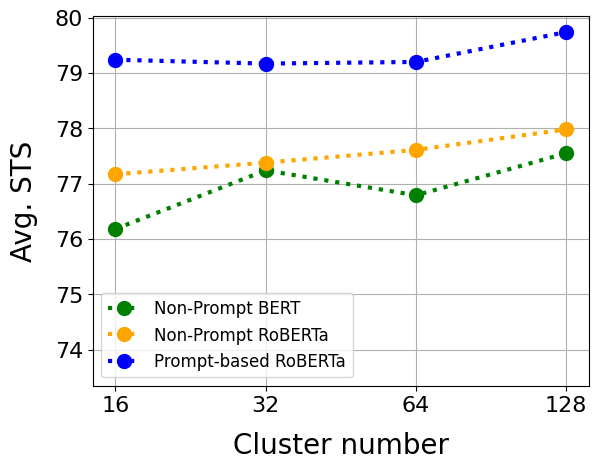}
    \caption{Comparisons of different cluster number.}
    \label{fig:cluster-k}
\end{figure}

We use large batch sizes and the cluster number $K$ for our models in the main experiments. To show the necessity, we implement the quantitative analysis to compare with small batch sizes and cluster numbers, and show the results in Figure~\ref{fig:bsz} and Figure~\ref{fig:cluster-k}. Both experiments of small batch sizes and cluster numbers perform worse. We attribute the performance degeneration to three factors: 1) Contrastive learning requires large batch sizes in general; 2) Smaller cluster numbers lead to more coarse-grained clusters, weakening the clustering performance; And 3) small batch sizes further restrain the number of clusters.

\subsection{Centroids Initialization}
\begin{figure}[htbp]
    \centering
    \includegraphics[width=0.9\linewidth]{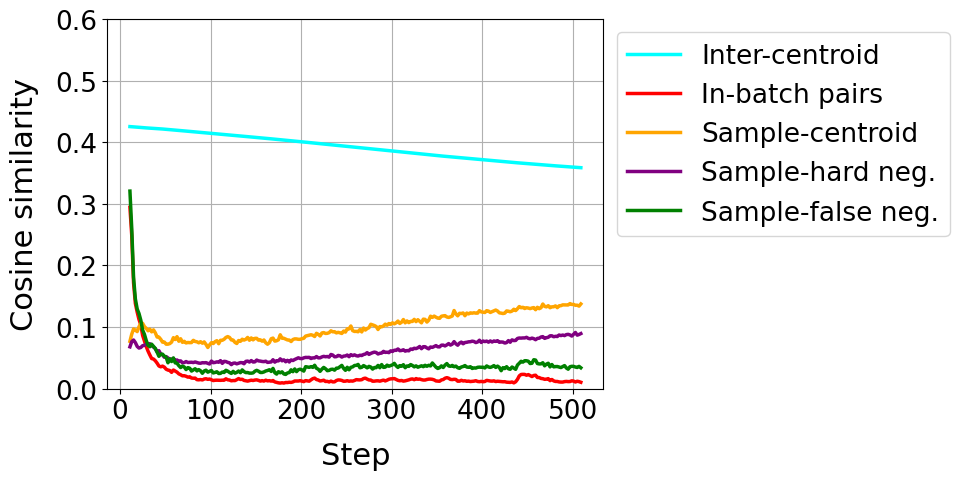}
    \caption{Variation for similarity of sample-nearest centroid pairs (Sample-centroid), sample-hard negative pairs, inter-centroid pairs, intra-cluster member pairs and in-batch negative pairs in global ClusterNS, corresponding to Figure~\protect\ref{fig:cluster}.}
    \label{fig:global-init}
\end{figure}

We initialize the cluster centroids locally as mentioned in Section~\ref{sec:K-means method}. While some other works adopt global initialization \cite{li2020prototypical}, they take the embeddings of whole dataset to initialize the centroids. We compare the two strategies by implementing the global initialization version of ClusterNS (named global ClusterNS). We show the test results in Table~\ref{tab:cluster-init}, and the variation of clustering similarity in Figure~\ref{fig:global-init}. Overall, global ClusterNS does not improve the performance. We obverse that inter-centroid pairs have extreme high similarity, meaning that clusters do not scatter, and the similarity of hard negative pairs is very low, which means hard negatives are not able to provide strong gradient signal.

\begin{table}[htbp]
\centering
\resizebox{1\linewidth}{!}{
\begin{tabular}{lc}
\hline
Models & Avg. STS \\ \hline
ClusterNS-$\mathrm{RoBERTa_{base}}$ & 77.98 \\
Global ClusterNS-$\mathrm{RoBERTa_{base}}$ & 77.81 \\ \hline
\end{tabular}
}
\caption{Comparsion of different centroid initialization methods with \emph{Non-Prompt} ClusterNS-$\mathrm{RoBERTa_{base}}$.}
\label{tab:cluster-init}
\end{table}

\begin{table*}[btp]
\centering
\renewcommand\arraystretch{1.1}
\resizebox{1\linewidth}{!}{
\begin{tabular}{l}
\hline
Example 1 \\ \hline
\#1: Jantroon as a word is derived from an Urdu word [UNK] which means Paradise. \\
\makecell[l]{\#2: While the liturgical atmosphere changes from sorrow to joy at this service, the faithful continue to fast \\and the Paschal greeting, "Christ is risen!} \\
\#3: There is also a Methodist church and several small evangelical churches. \\
\#4: Hindu Temple of Siouxland \\
\#5: Eventually, the original marble gravestones had deteriorated, and the cemetery had become an eyesore. \\
\makecell[l]{\#6: Reverend Frederick A. Cullen, pastor of Salem Methodist Episcopal Church, Harlem's largest \\congregation, and his wife, the former Carolyn Belle Mitchell, adopted the 15-year-old Countee Porter, \\although it may not have been official.} \\
\#7: The also include images of saints such as Saint Lawrence or Radegund. \\ \hline
Example 2 \\ \hline
\makecell[l]{\#1: Besides Bach, the trio recorded interpretations of compositions by Handel, Scarlatti, Vivaldi, Mozart, \\ Beethoven, Chopin, Satie, Debussy, Ravel, and Schumann.} \\
\#2: Guitarist Jaxon has been credited for encouraging a heavier, hardcore punk-influenced musical style. \\
\makecell[l]{\#3: Thus, in Arabic emphasis is synonymous with a secondary articulation involving retraction of the \\ dorsum or root of the tongue, which has variously been} \\
\#4: MP from January, 2001 to date. \\
\#5: The song ranked No. \\
\makecell[l]{\#6: The tones originate from Brown's acoustic Martin guitar, which is set up through two preamplifiers \\ which are connected to their own power amplifiers.} \\ \hline
\end{tabular}
}
\caption{Illustrative examples in clusters resulting from ClusterNS. Sentences with similar topics are grouped into clusters.}
\label{tab:cluster-example}
\end{table*}

\end{document}